\pdfoutput=1

\documentclass[11pt]{article}

\usepackage[preprint]{acl}

\usepackage{times}
\usepackage{latexsym}
\usepackage{amssymb}
\usepackage[T1]{fontenc}

\usepackage[utf8]{inputenc}

\usepackage{microtype}

\usepackage{inconsolata}

\usepackage{graphicx}
\usepackage{amsmath}
\usepackage{multirow}
\usepackage{booktabs} 
\usepackage{makecell, hhline}
\usepackage{tabularx, threeparttable, tabulary}
\usepackage{siunitx} 
\usepackage{algorithm}
\usepackage{algpseudocode}
\usepackage{adjustbox}
\usepackage{xcolor}
\usepackage{listings}
\usepackage{subcaption}   
\definecolor{lightred}{RGB}{255, 102, 102}
\definecolor{lightblue}{RGB}{102, 178, 255}
\title{Coordinating Search-Informed Reasoning and Reasoning-Guided Search in Claim Verification}

\author{Qisheng Hu \quad Quanyu Long \quad Wenya Wang \\
        Nanyang Technological University \\
        \texttt{qisheng001@e.ntu.edu.sg}
        }

\begin{document}
\maketitle
\begin{abstract}
Multi-hop claim verification is inherently challenging, requiring multi-step reasoning to construct verification chains while iteratively searching for information to uncover hidden bridging facts. This process is fundamentally interleaved, as effective reasoning relies on dynamically retrieved evidence, while effective search demands reasoning to refine queries based on partial information. To achieve this, we propose \textbf{H}ierarchical \textbf{A}gent \textbf{R}easoning and \textbf{I}nformation \textbf{S}earch (HARIS), explicitly modeling the coordinated process of reasoning-driven searching and search-informed reasoning. HARIS consists of a high-level reasoning agent that focuses on constructing the main verification chain, generating factual questions when more information is needed, and a low-level search agent that iteratively retrieves more information, refining its search based on intermediate findings. This design allows each agent to specialize in its respective task, enhancing verification accuracy and interpretability. HARIS is trained using reinforcement learning with outcome-based rewards. Experimental results on the EX-FEVER and HOVER benchmarks demonstrate that HARIS achieves strong performance, greatly advancing multi-hop claim verification. \footnote{Code: \url{https://github.com/qishenghu/HARIS}}
\end{abstract}

\section{Introduction}
Claim verification~\citep{guo2022survey} has become a critical challenge as misinformation proliferates online. It requires systems to determine whether a given claim is supported or refuted based on retrieved evidence. While verifying simple claims involves shallow reasoning within a single document, the verification of complex, multi-hop claims presents a fundamentally different challenge. This difficulty stems from the fragmented nature of evidence~\citep{pham2025verify, atanasova2022fact}. Effectively verifying such claims requires a joint process of multi-step reasoning and iterative information searching~\citep{zheng2025mrr}.

\begin{figure}[t]
    \centering
    \includegraphics[width=1.0\linewidth]{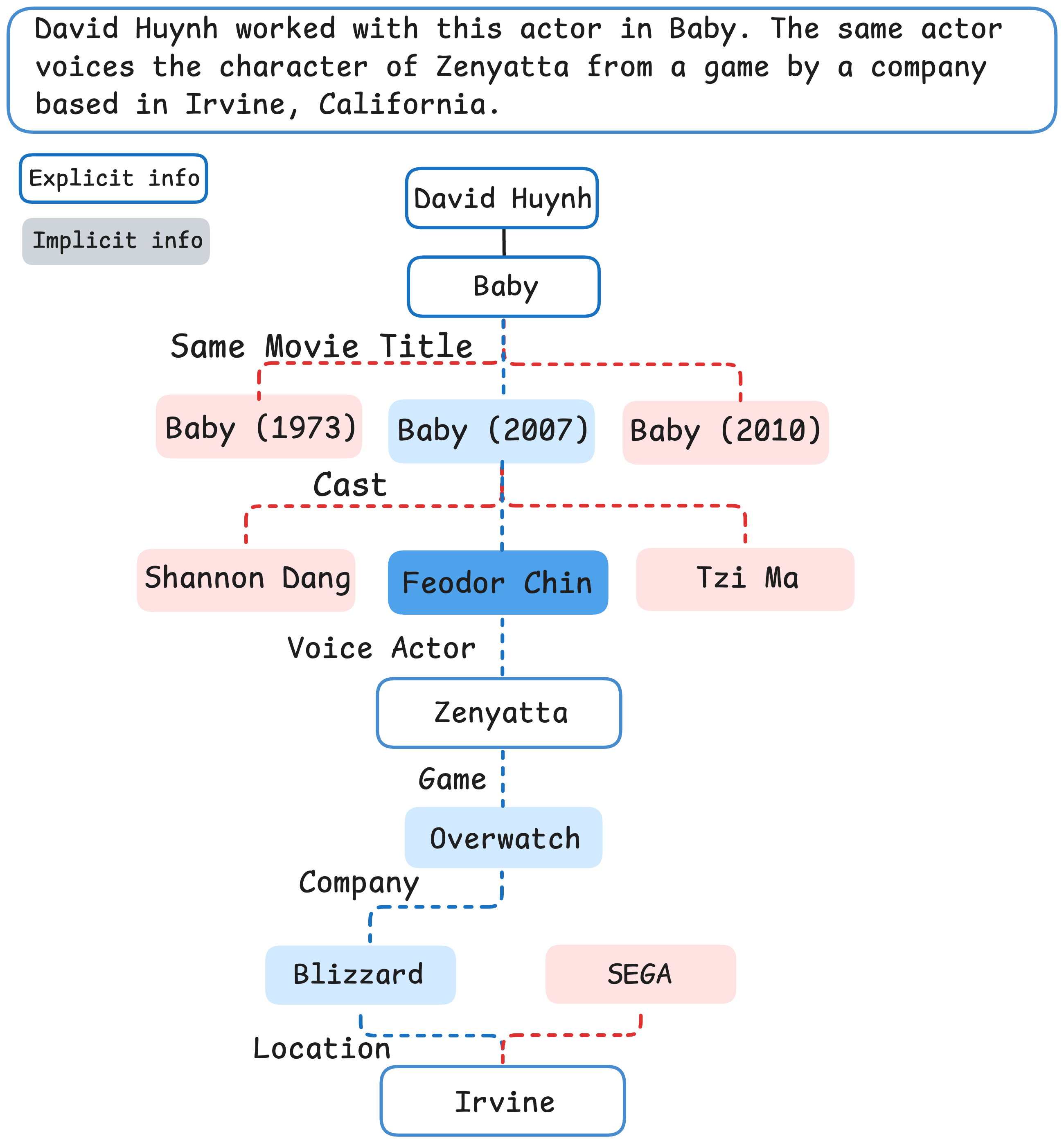}
    \caption{Example of a challenging multi-hop verification. Verifying the claim demands coordinating reasoning-guided search to disambiguate entities and search-informed reasoning to adapt based on retrieved evidence. Prematurely concluding on any distracting branch (in red) leads to incorrect judgment. The correct path—from \emph{Baby (2007)} to \emph{Overwatch} via \emph{Feodor Chin}—emerges only through this dynamic interplay, not static decomposition or single-pass retrieval.}
    \label{fig:claim_example}
\end{figure}

A key challenge in multi-hop verification is identifying the correct bridging facts—implicit links that connect separate pieces of evidence but are not explicitly stated. As shown in Figure~\ref{fig:claim_example}, verifying this claim requires identifying \emph{Feodor Chin} as the critical bridging fact between the film \emph{Baby (2007)} and the character \emph{Zenyatta} from the game \emph{Overwatch}, among other irrelevant paths. This process necessitates intensive interactions between \textbf{reasoning}, which proposes candidate hypotheses to identify potential bridging facts, and \textbf{iterative search}, which retrieves evidence to validate or eliminate certain hypotheses. In particular, reasoning is necessary to construct verification chains, but effective reasoning depends on relevant evidence, which often requires iterative search. Meanwhile, effective search relies on reasoning to formulate queries based on partially retrieved evidence. This reciprocal relationship, where reasoning shapes search and retrieved evidence refines ongoing reasoning, captures the recursive nature of multi-hop verification.

Conventional approaches typically involve decomposing complex claims into sub-claims or questions, followed by independent verification~\citep{kamoi-etal-2023-wice,chen-etal-2024-complex,lu2025optimizing}. However, this strategy can struggle when critical bridging facts are implicit and not directly recoverable from the claim. More advanced methods impose structured reasoning frameworks, such as graphs, reasoning programs, or First-Order Logic (FOL), to better coordinate evidence collection and reasoning~\citep{pan-etal-2023-fact,wang-shu-2023-explainable,pham2025verify}. However, these approaches often overlook the dynamic interplay between reasoning and information retrieval, which can be critical for accurate multi-hop verification.

To tackle these, we propose HARIS—\textbf{H}ierarchical \textbf{A}gent \textbf{R}easoning and \textbf{I}nformation \textbf{S}earch, explicitly modeling the coordinated process of reasoning-driven searching and search-informed reasoning. HARIS consists of two specialized large language model (LLM) agents: a high-level reasoning agent and a low-level search agent, both trained using reinforcement learning (RL) to optimize their respective tasks. The high-level agent forms the main verification chain, generating factual questions when more information is needed. The low-level agent handles these questions through dynamic search, iteratively refining its queries based on partial results to progressively build a comprehensive evidence base. This design allows each agent to specialize in its respective task, enhancing both verification accuracy and interpretability by clearly modeling the mutually reinforcing interaction between reasoning and information searching. To our knowledge, HARIS is among the first RL-based cooperative agent approach for claim verification.

Our contributions are summarized as follows:
\begin{itemize}
\item We propose HARIS, a hierarchical agent framework designed for complex multi-hop claim verification, explicitly modeling the coordinated process of reasoning-driven searching and search-informed reasoning.
\item HARIS is trained end-to-end via Group Relative Policy Optimization with outcome-based rewards, directly optimizing task performance without intermediate supervision. 
\item HARIS demonstrates strong performance on two challenging benchmarks, EX-FEVER and HOVER, validating its effectiveness in tackling multi-hop claim verification.
\end{itemize}

\begin{figure*}[ht]
\centering
\includegraphics[width=1.\textwidth]{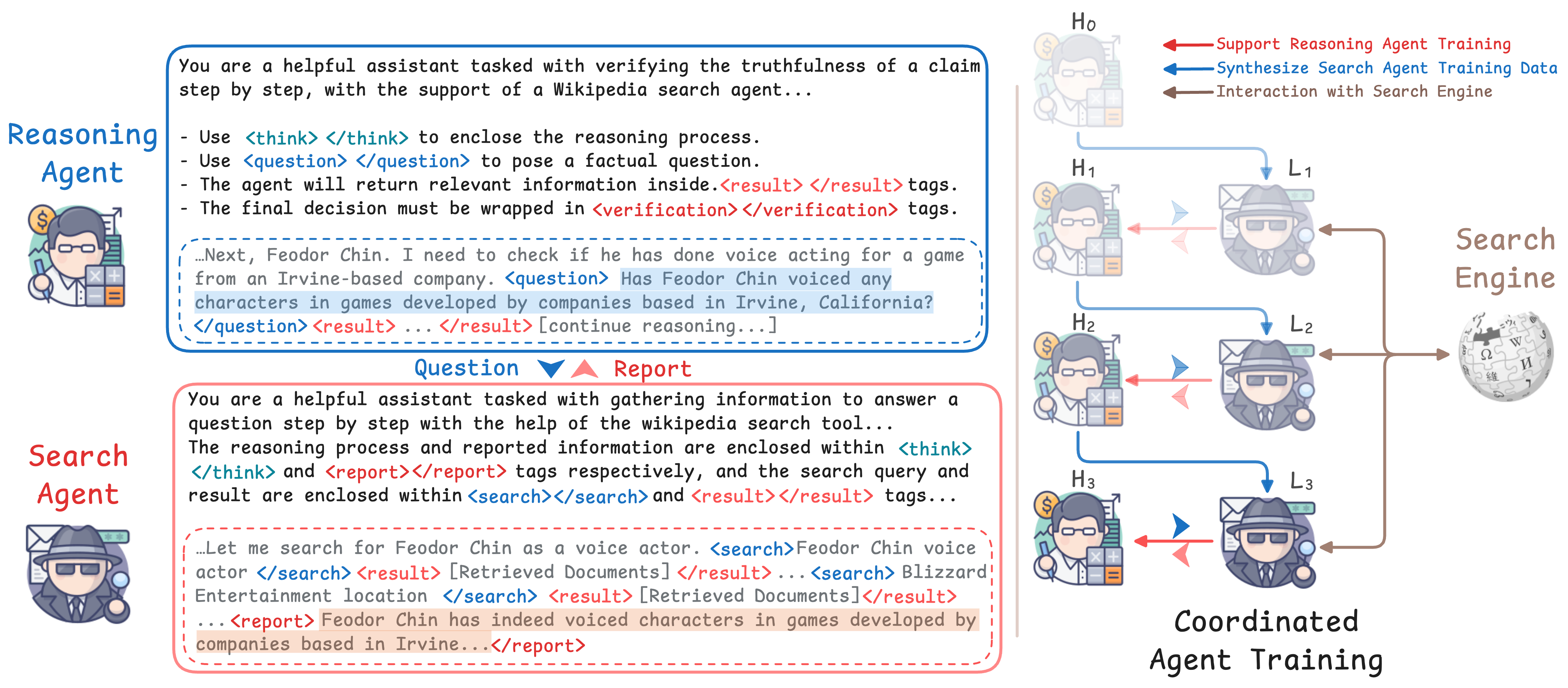}
\caption{Interaction workflow (left) and coordinated agent training process (right). The reasoning agent constructs the verification chain and issues questions (\texttt{<question>}) to the search agent. The search agent performs iterative retrievals (\texttt{<search>}) and report relevant information (\texttt{<report>}) back to the reasoning agent. During training, QA data from reasoning rollouts is used to update the search agent, which in turn supports reasoning agent updates, keeping both agents aligned.} 
\label{fig:bootstrap_workflow}
\end{figure*}

\section{Related Work}
\subsection{Claim Verification}
Claim verification research has been increasingly focusing on enhancing transparency during decision making~\citep{zeng-gao-2024-justilm,chen-etal-2024-complex, hu2025boost}. QACheck~\citep{pan-etal-2023-qacheck} reformulates verification as a progressive question-answering (QA) task, validating claims through step-wise questioning. Structural approaches~\citep{jeon2025graphcheck, pham2025verify}, such as ProgramFC~\citep{pan-etal-2023-fact} and FOLK~\citep{wang-shu-2023-explainable}, which use symbolic reasoning or reasoning program to enforce systematic verification. The Decompose-Then-Verify~\citep{wanner2024dndscore,wanner-etal-2024-closer,jiang2024core,hu-etal-2025-decomposition,lu2025optimizing} paradigm focus on breaking down complex claims into simpler sub-claims for independent validation. Agentic approaches have emerged as a promising direction~\citep{zhao2024pacar}. LoCal~\citep{ma2025local} employs a prompt-driven multi-agent framework emphasizing causal consistency, and BiDeV~\citep{liu2025bidev} uses specialized agents to address vagueness and redundancy. In contrast, HARIS formulates claim verification as a cooperative process between reasoning and search agents, and trains both agents jointly via reinforcement learning.

\subsection{Reasoning \& Searching}
Recent work has expanded the reasoning capabilities of LLMs by integrating search mechanisms~\citep{xiong2025rag, guan2025deeprag, sun2025zerosearch}. Notable methods like Search-o1~\citep{li2025search} incorporate dynamic search into reasoning frameworks, improving factual accuracy in open-domain and multi-hop reasoning tasks leveraging LLMs. Furthermore, RL methods like Group Relative Policy Optimization (GRPO)~\citep{shao2024deepseekmath} have been used to incentivize search capabilities of LLMs~\citep{gao2025synergizing}, encouraging models to generate context-aware queries and effectively integrate retrieved information~\citep{jin2025search, song2025r1, chen2025research,qian2025toolrl,wu2025agentic}. Collectively, these approaches demonstrate that RL-trained search integration can effectively improve performance on knowledge-intensive tasks.

\section{Methodology}

\subsection{Why Reasoning and Search Agents?}
Verification reasoning and factual information searching demand fundamentally different capabilities. Reasoning requires multi-step planning, identifying hidden facts, and maintaining logical consistency across long contexts. In contrast, effective searching depends on precise query formulation, iterative refinement, and robust extraction of relevant evidence from noisy or incomplete results. Rather than overloading a single model with both tasks, we decouple these roles into two specialized agents: a reasoning agent that interprets the claim, tracks verification progress, and decides when new information is needed; and a search agent that dynamically retrieves and refines evidence through focused interaction with a retrieval system.

Inspired by human cognition, where individuals offload information gathering to collaborators to reduce burden, HARIS is designed to mimic this process. By delegating search and reasoning to distinct agents, we improve decision traceability in multi-hop claim verification. This design is especially effective in cases requiring nuanced disambiguation and step-wise evidence composition.

Meanwhile, the search and reasoning agents closely collaborate to enhance overall performance. Interactive generation allows the agents to iteratively guide each other's outputs. Coordinated training alternately optimizes the agents to strengthen their collaboration, as shown in Figure~\ref{fig:bootstrap_workflow}.


\subsection{High-level Reasoning Agent}
The high-level reasoning agent constructs the main verification chain, coordinating multi-step reasoning and generating factual questions for the search agent when additional information is needed.

\subsubsection{Reasoning Agent Rollout}
Following prior work~\citep{chen2025research, jin2025search}, the reasoning agent’s rollout process uses special tags to define question actions. Specifically, the tags \texttt{<question>} and \texttt{</question>} indicate that an action to call the search agent should be invoked. Upon detecting the \texttt{</question>} tag, the generation is paused and the enclosed content will be regarded as the factual question and sent to the search agent. The reported information from search agent is wrapped within \texttt{<result>} tags and appended to the sequence to enable continued rollout. The rollout ends when a final verification is derived, wrapped within \texttt{<verification>} tags. The prompt template is provided in Appendix~\ref{appendix:prompt_template}.

\subsubsection{Reasoning Agent Reward}
The reasoning agent is responsible for performing multi-step reasoning over the gathered evidence to verify the claim. It provides a binary verification decision, and the reward is the correctness of this decision. The overall reward $R_{\text{high}}$ is given by:

\begin{equation}
\small
R_{\text{high}} = 
\begin{cases} 
1, & \text{if correct prediction} \\
0.1, & \text{if wrong prediction, correct format} \\
0, & \text{if wrong format}
\end{cases}
\end{equation}

\subsection{Low-level Search Agent}
The low-level search agent in HARIS is responsible for handling factual questions generated by the reasoning agent. Unlike conventional question-answering systems that aim to produce concise answers, the search agent here is designed to iteratively gather comprehensive and relevant information to support the high-level reasoning process.

\subsubsection{Training Data Synthesis}
To train the search agent, it is crucial that the training data not only includes diverse question-answer pairs but also closely aligns with the type of the questions generated by the high-level agent.  This alignment ensures that the search agent can effectively collaborate with the reasoning agent during multi-hop verification. To create such training set, we sample questions from the high-level agent rollouts, pair them with the corresponding ground-truth evidence, and use GPT-4o~\citep{gpt4oreport} to generate pseudo ground-truth answers. To maintain data quality, we filter out pairs where GPT-4o outputs "none" as the answer, typically removing about 10\% of the data.  This process synthesize data that is both contextually relevant and closely matched to the reasoning patterns of the high-level agent. Synthesis details and examples can be found in Appendix~\ref{appendix:synthesis} and Table~\ref{tab:synthesized_data_examples}.

\subsubsection{Search Agent Rollout}
The search agent rollout process is similar to the reasoning agent. When the generation process encounters a \texttt{</search>} tag, it pauses and extracts the content wrapped within the tags as the search query. This query is then used to perform top-k retrieval from the knowledge corpus. The retrieved text is wrapped in \texttt{<result>} and \texttt{</result>} tags and appended to the paused sequence, allowing the rollout to continue iteratively until the agent determines that sufficient information has been gathered. At this point, the agent reports the collected evidence within \texttt{<report>} and \texttt{</report>} tags, marking the completion of the rollout. The prompt template can be found in Appendix~\ref{appendix:prompt_template}.

\subsubsection{Search Agent Reward}
\label{sec:search_agent_reward}
We use a combination of format and LLM-as-a-Judge approaches for search agent rewards. The LLM-as-a-Judge approach evaluates the quality the gathered information using an LLM, comparing the final output against the pseudo ground-truth answer\footnote{We use GPT-4o-mini\citep{gpt4ominireport} for this purpose, as it is cost-efficient.}. The evaluation score is computed as: 
\begin{equation}
S = \text{LLM-as-a-Judge}(a_{\text{pred}}, a_{\text{gt}})
\end{equation}
where ${a_\text{pred}}$ is the search agent's final output and ${a_\text{gt}}$ is the pseudo ground-truth answer. This approach is less strict than exact match (EM) metrics, better aligning with the search agent's goal of collecting comprehensive, contextually relevant information rather than just short, exact answers. The implementation can be found in Appendix~\ref{appendix:llm_as_a_judge}.  Formally, the reward $R_{\text{low}}$ is given by: 

\begin{equation}
\small
R_{\text{low}} = 
\begin{cases} 
1, & \text{if S > 0} \\
0.1, & \text{if S = 0, correct format} \\
0, & \text{if wrong format}
\end{cases}
\end{equation}

To assess the reliability of LLM-as-a-Judge, we recruited two annotators\footnote{\url{https://www.prolific.com/}} for assessing held-out samples using the same criteria as LLM. The results showed strong consistency with human judgment (Cohen’s Kappa: 0.81; agreement: 93.3\%). See Appendix~\ref{appendix:human_eval} for complete human evaluation details.

\begin{figure*}[ht]
\centering
\includegraphics[width=1.\textwidth]{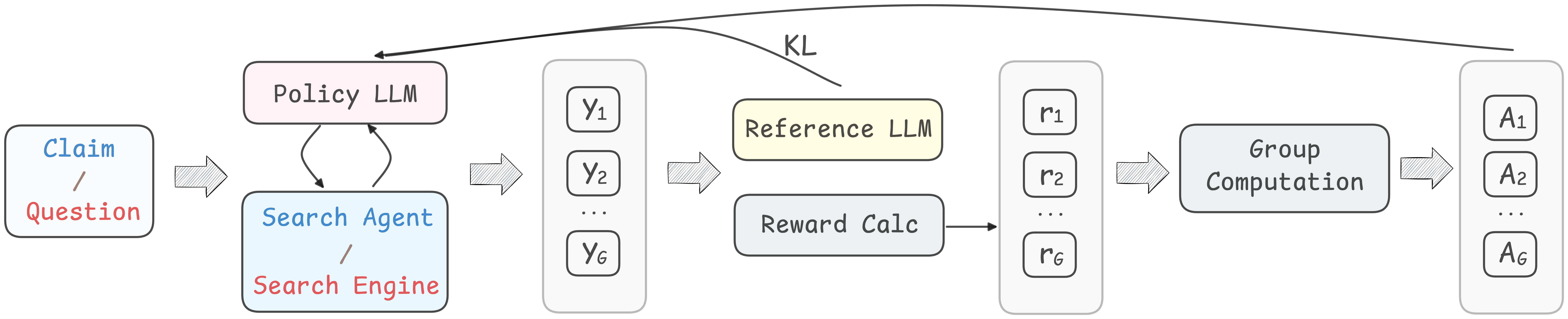}
\caption{Demonstrations of GRPO for the \textcolor{lightblue}{reasoning} and \textcolor{lightred}{search} agents in HARIS. During \textcolor{lightblue}{reasoning} agent training, the policy LLM interacts with the search agent, while for \textcolor{lightred}{search} agent training, it interacts with the search engine.}
\label{fig:GRPO}
\end{figure*}

\subsection{Group Relative Policy Optimization}
In this work, we leverage Group Relative Policy Optimization (GRPO)~\citep{shao2024deepseekmath}, an RL algorithm tailored for training LLMs with group-level reward normalization. GRPO introduces a relative advantage mechanism, which evaluates the quality of generated responses within groups corresponding to the same input. This design helps stabilize training by reducing variance in gradient updates, thereby promoting more consistent learning. The GRPO objective is formally defined as:

\begin{equation}
\small
\begin{aligned}
\mathcal{L}_{\text{GRPO}}(\theta) = -\frac{1}{G} \sum_{i=1}^G \sum_{t=1}^{|o_i|} \bigg[ 
    \frac{\pi_\theta(o_{i,t} \mid q, o_{i, <_t})}{\pi_{\theta_{\text{old}}}(o_{i,t} \mid q, o_{i, <_t})} \hat{A}_{i,t} \\
    - \beta D_{\text{KL}}[\pi_\theta \parallel \pi_{\theta_{\text{old}}}] 
\bigg],
\end{aligned}
\end{equation}

where $G$ is the number of groups, $\hat{A}_{i,t}$ is the normalized advantage within the group, and $\beta$ controls the KL divergence penalty enforcing policy stability. Figure \ref{fig:GRPO} presents an overview of GRPO for reasoning and search agents.

\paragraph{Retrieved Token Loss Masking} 
Following prior work~\citep{chen2025research,song2025r1}, retrieved tokens are masked during loss calculation. This approach ensures that the policy gradient is computed only on generated tokens, reducing bias toward retrieved content and stabilizing training.

\subsection{Coordinated Agent Training}
To enable effective collaboration between the high-level reasoning agent and the low-level search agent, we adopt a \textit{Coordinated Agent Training} strategy. This process consists of two stages: an initial \textbf{foundation stage}, where both agents develop core reasoning, search, and formatting abilities; and a subsequent \textbf{alternating training stage}, where the agents iteratively refine their specialized skills through mutual interaction.

Given a training set $T$ of claim verification data, in the foundation stage, the untrained low-level search agent $L_0$ and high-level reasoning agent $H_0$ are trained sequentially to establish their foundational capabilities. First, we sample questions from $H_0$ to train $L_0$, producing the updated search agent $L_1$. This updated search agent then supports the training of $H_0$, producing the updated reasoning agent $H_1$. This stage establishes a foundation for both agents in reasoning, searching, and formatting.

In the alternating stage, we promote further coordination by repeatedly alternating training between the two agents. The dataset $T$ is divided into $N$ segments. For each segment, we sample questions from the current high-level agent $H_i$ to train the low-level agent, producing $L_{i+1}$. The updated low-level agent $L_{i+1}$ is then used to collaborate with the high-level agent, resulting in $H_{i+1}$. This alternating process continues for $N$ rounds, fostering mutual adaptation and ensuring the agents remain closely aligned throughout training.


\begin{algorithm}[t]
\caption{Coordinated Agent Training}
\begin{algorithmic}
\State \textbf{Input:} Initial low-level agent $L_0$, high-level agent $H_0$, training set $T$
\State \textbf{Output:} Trained low-level agent $L$ and high-level agent $H$
\State \textbf{Stage 1: Foundation Training}
\State $Q_0 \gets \text{Synthesis}(T, H_0)$
\State $L_1 \gets GRPO_{low}(Q_0, L_0)$
\State $H_1 \gets GRPO_{high}(T, L_1)$
\State \textbf{Stage 2: Alternating Training}
\State Divide $T$ into $N$ segments $\{T_1, \ldots, T_N\}$
\For{$i = 1$ to $N$}
    \State $Q_i \gets \text{Synthesis}(T_i, H_i)$
    \State $L_{i+1} \gets GRPO_{low}(Q_i, L_i)$
    \State $H_{i+1} \gets GRPO_{high}(T_i, L_{i+1})$
\EndFor
\State $L \gets L_{N+1}$, $H \gets H_{N+1}$
\State \textbf{Return} $L, H$
\end{algorithmic}
\end{algorithm}

\begin{table*}[!ht]
    \centering
    \small
    \renewcommand\arraystretch{1.05}
    \setlength{\abovecaptionskip}{0.2cm}
    \setlength{\belowcaptionskip}{-0.3cm}
    \addtolength{\tabcolsep}{-0.5pt}
    \begin{threeparttable}
\begin{tabular}{@{}l
                S[table-format=2.2] 
                S[table-format=2.2]
                S[table-format=2.2]
                S[table-format=2.2]
                S[table-format=2.2]
                S[table-format=2.2]
                S[table-format=2.2]
                S[table-format=2.2]
                S[table-format=2.2]
                S[table-format=2.2]
                S[table-format=2.2]@{}}
\toprule
 &  \multicolumn{6}{c}{\textbf{HOVER}} & \multicolumn{4}{c}{\textbf{EX-FEVER}} & \textbf{CHECKWHY} \\
 &  \multicolumn{2}{c}{\textbf{2-hops}} & \multicolumn{2}{c}{\textbf{3-hops}} & 
\multicolumn{2}{c}{\textbf{4-hops}} & \multicolumn{2}{c}{\textbf{2-hops}} &  \multicolumn{2}{c}{\textbf{3-hops}} &  \\
 \cmidrule[0.4pt](r{0.125em}){2-3}%
 \cmidrule[0.4pt](r{0.125em}){4-5}%
 \cmidrule[0.4pt](r{0.125em}){6-7}%
 \cmidrule[0.4pt](r{0.125em}){8-9}%
 \cmidrule[0.4pt](r{0.125em}){10-11}%
 \cmidrule[0.4pt](lr{0.125em}){12-12}%
 & F\text{1} & Acc & F\text{1} & Acc & F\text{1} & Acc & F\text{1} & Acc & F\text{1} & Acc & Acc \\ 
\midrule
RAG & 58.96 & 59.20 & 56.59 & 56.60 & 55.06 & 55.20 & 68.83 & 69.00 & 64.41 & 64.80 & 56.40 \\
Decomp-Verify & 62.39 & 62.60 & 57.27 & 57.31 & 54.55 & 55.60 & 68.03 & 68.40 & 61.95 & 63.00 & 36.00\\
ProgramFC & 66.84 & 66.80 & 55.35 & 56.80 & 48.30 & 52.60 & 71.28 & 71.60 & 60.34 & 62.40 & 21.60 \\
BiDeV & 64.51 & 65.00 & 57.59 & 58.60 & 54.94 & 57.00 & 67.75 & 67.80 & 61.52 & 62.00 & 24.60 \\
QACheck & 67.60 & 67.60 & 60.60 & 60.60 & 58.91 & 59.00 & 75.52 & 75.60 & 68.42 & 68.60 & 58.00\\
FOLK & 67.22 & 67.60 & 59.89 & 61.20 & 50.90 & 55.20 & 75.55 & 75.80 & 67.24 & 68.40 & 50.70\\
Search-o1 & 68.72 & 69.00 & 59.34 & 59.80 & 54.90 & 56.60 & 77.41 & 77.80 & 72.08 & 72.80 & 52.40\\
\midrule
\textbf{HARIS} & \textbf{69.31} & \textbf{69.40} & \textbf{62.33} & \textbf{62.80} & \textbf{59.84} & \textbf{61.00} & \textbf{80.12} & \textbf{80.20} & \textbf{73.93} & \textbf{74.20} & \textbf{60.80} \\
\bottomrule
\end{tabular}
    \end{threeparttable}
\caption{Performance comparison of different methods on HOVER (2-hops/3-hops/4-hops), EX-FEVER (2-hops/3-hops) and CHECKWHY.}
\label{tab:main}
\end{table*}

\section{Experiments}

\subsection{Datasets}

We utilize the following datasets for training and evaluation:

\begin{itemize}
    \item \textbf{EX-FEVER}~\citep{ma-etal-2024-ex}: A benchmark for multi-hop claim verification, designed to assess a model’s ability to verify complex claims through 2-hop and 3-hop reasoning over hyperlinked Wikipedia documents.
    \item \textbf{HOVER}~\citep{jiang-etal-2020-hover}: A dataset created for many-hop claim verification, featuring claims that require 2 to 4-hop reasoning across multiple Wikipedia articles.
\end{itemize}

For training, we sample 7,200 examples from the combined EX-FEVER and HOVER training data. For evaluation, following \citet{wang-shu-2023-explainable}, we sample 500 instances from the test set of each dataset using stratified sampling, ensuring a balanced label distribution. We use F1-score and accuracy as the primary evaluation metrics. To further assess generalizability, we also evaluate the accuracy on 500 positive\footnote{CHECKWHY is a challenging benchmark. Its negative samples are created by modifying evidence to generate counterfactuals. Hence, we only use the positive samples.} test samples from CHECKWHY~\citep{si-etal-2024-checkwhy} in the main experiments. Details of the datasets can refer to Appendix~\ref{appendix:dataset}.

\subsection{Baselines}
We include the following baselines:
\paragraph{RAG} : A typical Retrieval-Augmented Generation (RAG) approach where retrieved documents and the input are provided to a LLM for verification. The verification module is implemented using DSPy~\citep{khattab2024dspy}.

\paragraph{Decompose-Then-Verify} A commonly used paradigm~\citep{kamoi-etal-2023-wice}  involving: decomposing a claim into sub-claims, verifying each independently, and aggregating the results. We utilize the decomposition module from~\citet{kamoi-etal-2023-wice} and prompt the LLM for verification, aggregating final results with logical \texttt{AND}.

\paragraph{ProgramFC} ~\citet{pan-etal-2023-fact} leveraged program-guided reasoning for claim verification, generating reasoning programs in a few-shot manner for execution.

\paragraph{QACheck} ~\citet{pan-etal-2023-qacheck} verifies claims through iterative question-answering until the LLM determines that sufficient information has been derived. We employ the default Retriever–Reader setting, where an LLM iteratively answers questions using the corpus.

\paragraph{FOLK} ~\citet{wang-shu-2023-explainable} translates claims into First-Order Logic (FOL) clauses and applies FOL-guided reasoning over knowledge-grounded question-answer (QA) pairs. The QA pairs are grounded via an external API\footnote{\url{https://serpapi.com/}}.

\paragraph{BiDeV} ~\citet{liu2025bidev} propose two prompt-based LLM agents for defusing vagueness and redundancy: the former clarifies latent information, while the latter removes redundant evidence.

\paragraph{Search-o1} ~\citet{li2025search} enhances large reasoning models by integrating agentic RAG, allowing autonomous retrieval during multi-step reasoning. It further refines retrieved information through a Reason-in-Documents module. 

For more baseline details, refer to Appendix~\ref{appendix:baselines}.

\begin{table}[t]
\centering
\small
\setlength{\abovecaptionskip}{0.2cm}
\setlength{\belowcaptionskip}{-0.3cm}
\begin{tabular}{lcccc}
\toprule
& \multicolumn{2}{c}{\textbf{Single}} & \multicolumn{2}{c}{\textbf{HARIS}} \\
\cmidrule(lr){2-3} \cmidrule(lr){4-5}
& F1 & Acc & F1 & Acc \\
\midrule
$\text{EX-FEVER}_{\text{2hops}}$ & 77.59 & 77.60 & 80.12 & 80.20 \\
$\text{EX-FEVER}_{\text{3hops}}$  & 73.20 & 73.40 & 73.93 & 74.20 \\
$\text{HOVER}_{\text{2hops}}$  & 68.20	& 68.20 & 69.31 & 69.40 \\
$\text{HOVER}_{\text{3hops}}$  & 61.55 & 62.00 & 62.33 & 62.80 \\
$\text{HOVER}_{\text{4hops}}$  & 55.48 & 56.00 & 59.84 & 61.00 \\
\bottomrule
\end{tabular}
\caption{Performance comparison between RL trained single agent and HARIS.}
\label{tab:agent_comparison}
\end{table}

\begin{table}[t]
\centering
\small
\setlength{\abovecaptionskip}{0.2cm}
\setlength{\belowcaptionskip}{-0.3cm}
\begin{tabular}{lcccc}
\toprule
\multirow{2}{*}{} & \multicolumn{2}{c}{\textbf{$N$=1}} & \multicolumn{2}{c}{\textbf{$N$=3}} \\
\cmidrule(lr){2-3} \cmidrule(lr){4-5}
& F1 & Acc & F1 & Acc \\
\midrule
$\text{EX-FEVER}_{\text{2hops}}$ & 78.77 & 78.80 & 80.12 & 80.20 \\
$\text{EX-FEVER}_{\text{3hops}}$ & 73.20 & 73.60 & 73.93 & 74.20 \\
$\text{HOVER}_{\text{2hops}}$ & 69.37 & 69.40 & 69.31 & 69.40 \\
$\text{HOVER}_{\text{3hops}}$ & 61.08 & 61.60 & 62.33 & 62.80 \\
$\text{HOVER}_{\text{4hops}}$ & 60.87 & 61.80 & 59.84 & 61.00 \\
\bottomrule
\end{tabular}
\caption{Comparison of performance for coordination training rounds $N$.}
\label{tab:round_comparison}
\end{table}

\begin{table}[ht]
\centering
\small
\setlength{\abovecaptionskip}{0.2cm}
\setlength{\belowcaptionskip}{-0.3cm}
\begin{tabular}{lcccc}
\toprule
& \multicolumn{2}{c}{\textbf{F1}} & \multicolumn{2}{c}{\textbf{LLM-as-a-Judge}} \\
\cmidrule(lr){2-3} \cmidrule(lr){4-5}
& F1 & Acc & F1 & Acc \\
\midrule
$\text{EX-FEVER}_{\text{2hops}}$ & 72.44 & 73.00 & 75.63 & 75.80 \\
$\text{EX-FEVER}_{\text{3hops}}$  & 63.85 & 65.60 & 68.14 & 68.80 \\
$\text{HOVER}_{\text{2hops}}$  & 65.94	& 66.40 & 67.49 & 67.60 \\
$\text{HOVER}_{\text{3hops}}$  & 56.42 & 57.80 & 58.97 & 59.60 \\
$\text{HOVER}_{\text{4hops}}$  & 52.38 & 55.60 & 57.42 & 58.80 \\
\bottomrule
\end{tabular}
\caption{Final performance comparison between HARIS with F1-trained and LLM-as-a-Judge trained search agent.}
\label{tab:reward_comparison}
\end{table}

\subsection{Experimental Setup}
We conduct all training experiments using the Qwen3-4B model~\citep{qwen3report}. For HARIS, we train for one epoch in each stage. All baseline methods, except for Search-o1, utilize GPT-4o~\citep{gpt4oreport} as the underlying LLM. Search-o1 employs the QwQ-32B-preview model~\citep{qwq-32b-preview}. All methods operate in the Open-Book setting~\citep{pan-etal-2023-fact}, where no ground-truth evidence is provided beforehand, requiring each method to retrieve supporting evidence using top-$k$ ($k=3$) retrieval. We utilize the Wikipedia corpus provided by FlashRAG~\citep{FlashRAG} for this purpose, indexed using a E5-small model for dense retrieval. For more training and experimental details, please refer to Appendix~\ref{appendix:training_settings}.

\section{Result}
\subsection{Main Result}

As shown in Tables~\ref{tab:main}, our proposed method HARIS consistently outperforms all baseline methods across both the EX-FEVER and HOVER datasets, demonstrating superior multi-hop reasoning and evidence searching capabilities. Notably, HARIS achieves the highest F1 and accuracy scores across different hop counts, with particularly strong performance in the more challenging 3-hop and 4-hop settings. For example, on the HOVER dataset, HARIS achieves 62.80\% accuracy in the 3-hop setting and 61.00\% in the 4-hop setting, surpassing other strong GPT-4o-powered baselines. For direct comparison with Qwen3-4B results, see Table~\ref{tab:qwen3_main}.

\begin{figure}[t]
\centering
\setlength{\abovecaptionskip}{0.2cm}
\setlength{\belowcaptionskip}{-0.3cm}
    \includegraphics[width=0.5\textwidth]{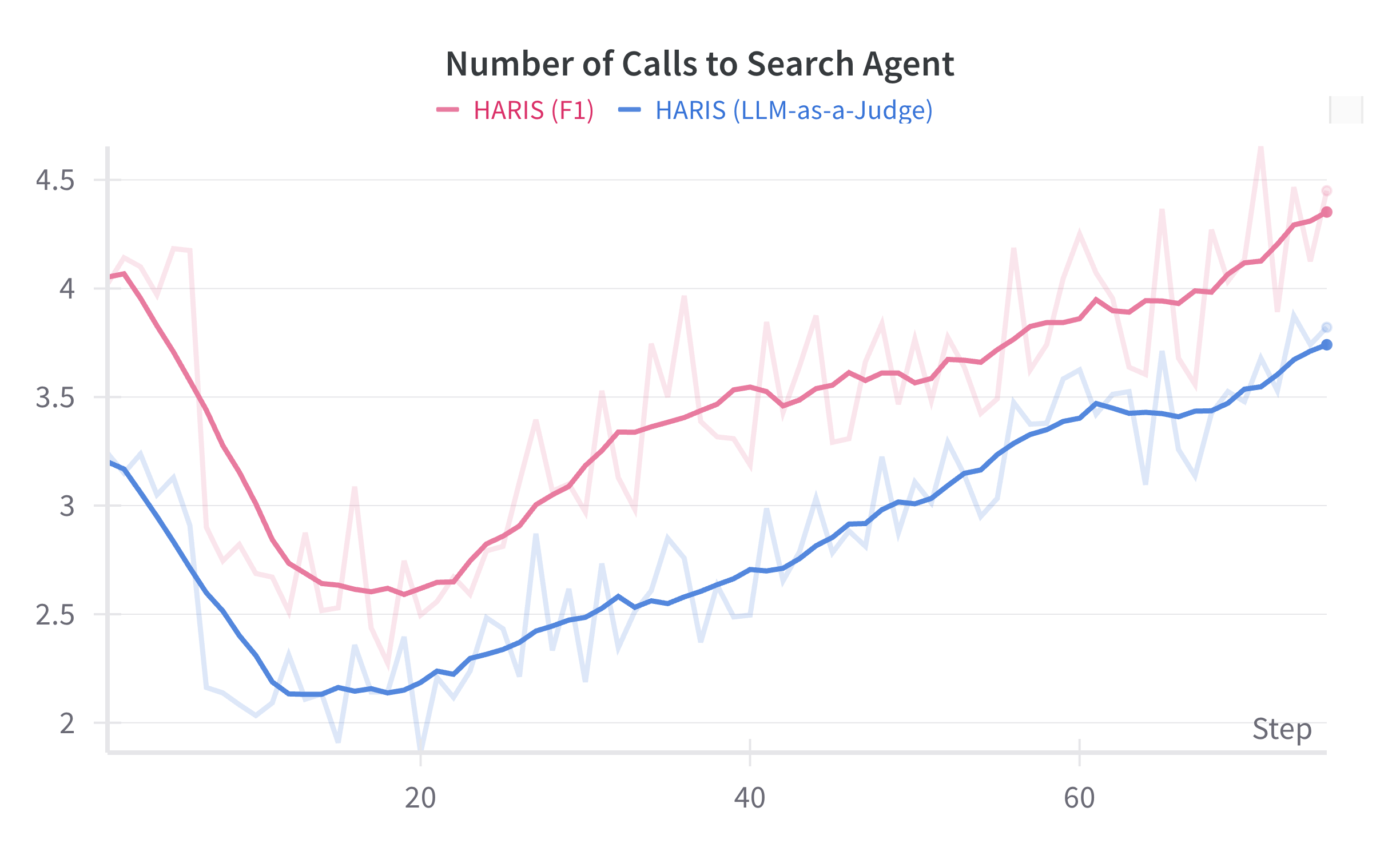}
    \caption{Comparision of calls to search agent during reasoning agent training, using search agents trained with F1 versus LLM-as-a-Judge rewards.}
    \label{fig:low_reward_comparison_search_cnt}
\end{figure}

\begin{figure}[t]
    \centering
\setlength{\abovecaptionskip}{0.2cm}
\setlength{\belowcaptionskip}{-0.3cm}
    \includegraphics[width=1.0\linewidth]{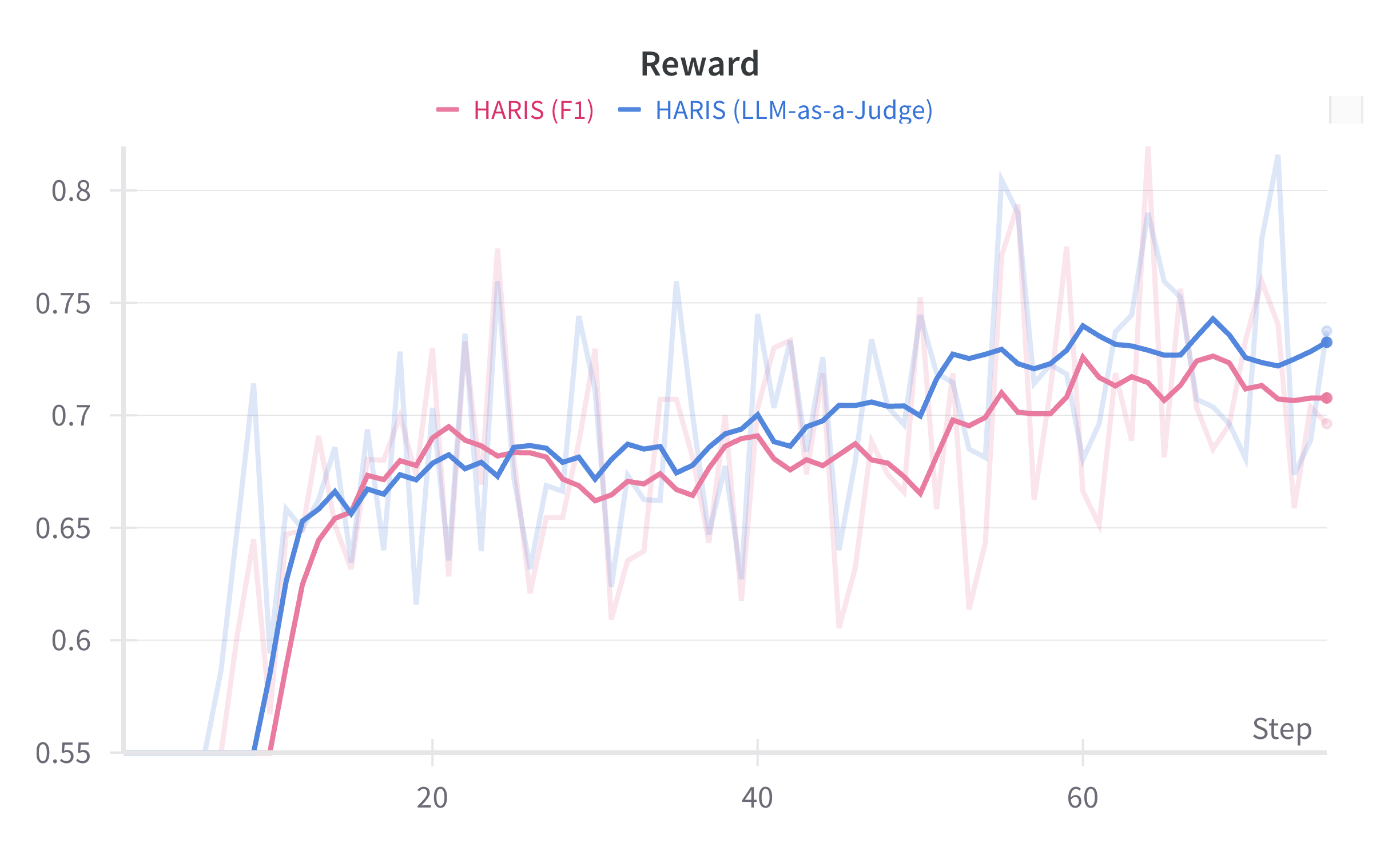}
    \caption{Comparison of high-level rewards during reasoning agent training with search agents trained using F1 and LLM-as-a-Judge rewards.}
    \label{fig:low_reward_comparison_reward}
\end{figure}

Notably, HARIS demonstrates strong generalization capabilities, achieving the best performance on the CHECKWHY benchmark. This result indicates that HARIS effectively handles more complex, causally structured claims, where gathering sufficient evidence and orchestrating multi-step reasoning is critical. This performance can be attributed to the explicit modeling of reasoning-driven searching and search-informed reasoning, which allows HARIS to dynamically refine verification paths based on partial evidence, reducing noise and improving verification consistency.

\begin{figure*}[ht]
\centering
\includegraphics[width=1.\textwidth]{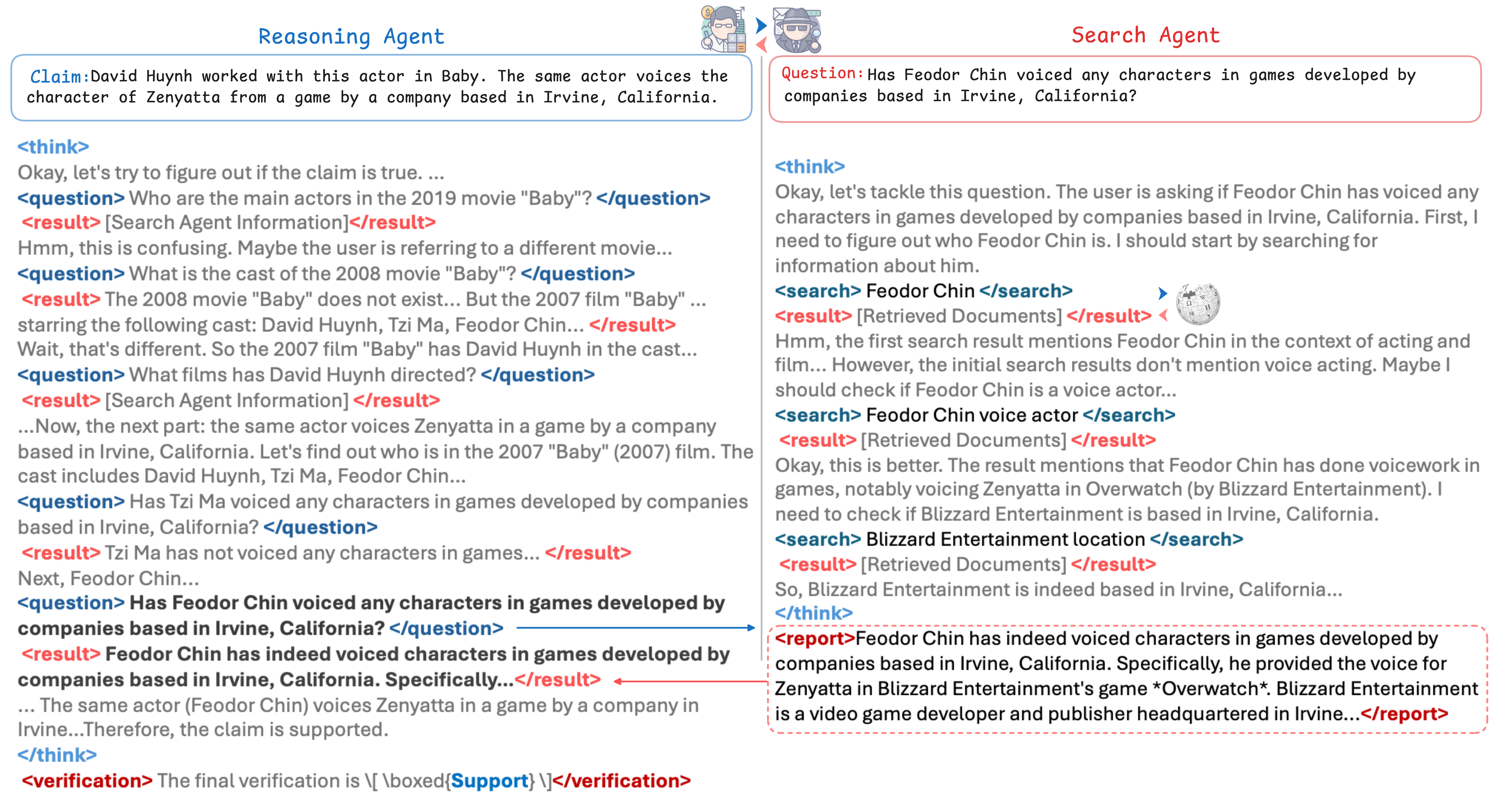}
\caption{Example of reasoning agent and search agent rollout interaction for a complex multi-hop claim. } 
\label{fig:case_study}
\end{figure*}

\subsection{Ablations}
\label{sec:ablations}
\paragraph{Single-Agent vs Multi-Agent}
Multi-agent RL has shown strong performance in complex reasoning tasks~\citep{wan2025rema}, and reducing retrieval noise during intermediate steps is also known to benefit RAG systems like Search-o1~\citep{li2025search}. To assess the impact of our bi-level design, we compare HARIS with an RL-trained single-agent . As shown in Table~\ref{tab:agent_comparison}, HARIS outperforms the single-agent setup across multiple datasets. This shows the advantage of decomposing reasoning and retrieval into specialized agents, each optimized for its specific role. The search agent efficiently provides relevant information to the reasoning agent, reducing noise and enhancing the verification capability. A qualitative case study in Appendix~\ref{appendix:single_case_study} further illustrates this contrast on a shared example.

\paragraph{Coordination Rounds}
We study how the number of coordination rounds ($N$) affects model performance. As shown in Table~\ref{tab:round_comparison}, increasing $N$ generally leads to stronger performance. It shows that this approach helps balance learning dynamics and maintain alignment between high-level and low-level agents. By allowing each agent to iteratively refine its abilities while maintaining consistency, coordinated training supports more effective collaboration over multiple training cycles.

\paragraph{LLM-as-a-Judge vs F1}
We examine the impact of the reward metric used for training the search agent using 3,600 training examples over one epoch. Specifically, we compare conventional QA F1-score and LLM-as-a-Judge as rewards. We find that F1-trained agents tend to generate more concise responses, as F1 favors answers closely matching the reference. In contrast, LLM-as-a-Judge rewards encourage more comprehensive and contextually relevant outputs. As shown in Figure~\ref{fig:low_reward_comparison_search_cnt} and ~\ref{fig:low_reward_comparison_reward}, F1-based agents prompt the reasoning agent to trigger more searches, resulting in more follow-up questions. In comparison, LLM-as-a-Judge reduces search calls but achieves higher verification reward, indicating more thorough information improves the overall reasoning process. As shown in Table~\ref{tab:reward_comparison}, the performance results suggest using LLM-as-a-Judge trained HARIS consistently improved the performance compared to F1-trained. On average it improves over 3\% performance.

More experiments can be found in Appendix~\ref{appendix:add_exp}.


\section{Case Study}
Figure~\ref{fig:case_study} illustrates how HARIS resolves a complex claim through step-by-step, search-informed reasoning. The reasoning agent systematically probes plausible related actors, while the search agent continuously refines queries, shifting the search from \emph{Feodor Chin} to \emph{Blizzard} to gather sufficient evidence. In another example (Figure~\ref{fig:case_study_2}), the reasoning agent initially struggles to identify the correct \emph{Baby} film, but the search agent’s response about "Who is David Huynh..." provides crucial context, steering reasoning toward the correct verification. These cases highlight HARIS’s collaborative process, with the reasoning agent refining its understanding as new information is retrieved until all critical connections are uncovered.

\section{Conclusion}
 We propose Hierarchical Agent Reasoning and Information Search (HARIS), explicitly modeling the coordinated process of reasoning-driven searching and search-informed reasoning. By integrating high-level reasoning and low-level search agents, HARIS effectively captures complex reasoning chains while reducing noise in evidence retrieval. Our approach demonstrates strong performance across challenging benchmarks, highlighting its effectiveness for comprehensive claim verification. 

\section*{Limitations}
While HARIS demonstrates strong performance, due to limited computational resources, we train only on a 4B model. Using larger models are likely to achieve even stronger performance. Additionally, our study focuses on binary claim verification (`Support' or `Refute'). While some benchmarks include additional classes such as `Neutral' or `Not Enough Info,' we do not explore them here. Claim verification is a key area within fact-checking, we do not explore tasks such as open-domain QA or counterfactual detection, as these differ substantially from multi-hop claim verification. Notably, our binary setting is consistent with strong baselines such as ProgramFC, QACheck, and FOLK in claim verification.



\bibliography{custom}

\appendix
\clearpage
\label{sec:appendix}

\section{Experimental Settings}
\subsection{Dataset}
\label{appendix:dataset}
HOVER\citep{jiang-etal-2020-hover} is a multi-hop claim verification dataset containing 2, 3, and 4-hop data, designed to evaluate the ability of models to connect multiple pieces of evidence across different documents. EX-FEVER\citep{ma-etal-2024-ex} is another multi-hop benchmark, primarily focused on 2 and 3-hop reasoning over hyperlinked Wikipedia documents.

For training, we sample 7,200 examples from the combined training sets of EX-FEVER and HOVER, maintaining an equal ratio across different hop lengths to ensure balanced coverage of multi-hop reasoning.

For evaluation, we follow ~\citet{wang-shu-2023-explainable} and use stratified sampling to select 500 instances for each hop setting, ensuring a balanced distribution of multi-hop complexity.

We also evaluate on CHECKWHY~\citep{si-etal-2024-checkwhy}, a challenging claim verification dataset where negative samples are constructed by modifying evidence to create counterfactuals. Given this design, we only sample from the positive claims and use accuracy for evaluation.

\subsection{Retrieval Setting}
We use the Wikipedia corpus processed by FlashRAG~\citep{FlashRAG}, which provide chunked passages. We adopt a dense retrieval method with `intfloat/multilingual-e5-small' model~\citep{wang2024multilingual}, which offers a favorable balance between memory efficiency and performance on the MTEB benchmark~\citep{muennighoff-etal-2023-mteb}. For retrieval, we use a top-3 retrieval strategy.

\subsection{Baselines}
\label{appendix:baselines}
For consistency, we adapt each baseline to the same experimental setup wherever possible.

\paragraph{RAG}
We use the input claim as the retrieval query, providing the retrieved context and the claim to GPT-4o for final classification. The verification signature is provided in Table~\ref{tab:signature}.

\paragraph{Decompose-Then-Verify}
We use the decomposition module from WICE~\citep{kamoi-etal-2023-wice} for breaking down complex claims into simpler sub-claims via few-shot in-context learning. Each sub-claim is then verified using the same retrieval and classification setup as RAG, with final results aggregated using logical AND.

\paragraph{ProgramFC}~\citep{pan-etal-2023-fact}:
We implement ProgramFC based on the official repository\footnote{\url{https://github.com/mbzuai-nlp/ProgramFC}}. To ensure consistency, we replace the Flan-T5 model used for sub-task functions with GPT-4o.

\paragraph{Search-o1}~\citep{li2025search}
Our implementation is based on the official implementation\footnote{\url{https://search-o1.github.io/}} and setting the maximum search limit to 10. We adapt the original QA prompt templates for claim verification.

\paragraph{FOLK}~\citep{wang-shu-2023-explainable}
Our implementation follows the official repository\footnote{\url{https://github.com/wang2226/FOLK}}. Consistent with the original paper, we perform knowledge grounding using the Google Search API\footnote{\url{https://serpapi.com/}}, ensuring accurate grounding for FOL-guided reasoning.

\paragraph{QACheck}~\citep{pan-etal-2023-qacheck}
We use the official implementation\footnote{\url{https://github.com/XinyuanLu00/QACheck}}. We replace the original LLM components with GPT-4o to match our baseline settings and ensure consistent evaluation.

\subsection{Search Agent Training Data Synthesis}
\label{appendix:synthesis}
To ensure the search agent can effectively address questions generated by the reasoning agent, we synthesize training data by having the reasoning agent perform rollouts on the training claims and sampling the generated questions.

For the first epoch training, we collect the first question proposed by the untrained reasoning agent $H_0$ in each rollout. This is because the initial un-trained reasoning agent struggles with formatting, making longer rollouts less reliable. Starting from the second epoch, we sample from all questions generated during the rollout as the reasoning agent at this stage has developed a more stable question generation capability. For training efficiency, in postprocessing, we limit each claim verification data to a single question.

To generate answers for these sampled questions, we pair each question with the ground-truth evidence provided by the original dataset. For EX-FEVER, we use the human-annotated explanations as the evidence. The prompt signature used for this pairing is provided in Table~\ref{tab:signature}.

\begin{table*}[!ht]
    \centering
    \small
    \renewcommand\arraystretch{1.05}
    \setlength{\abovecaptionskip}{0.2cm}
    \setlength{\belowcaptionskip}{-0.3cm}
    \addtolength{\tabcolsep}{-0.5pt}
    \begin{threeparttable}
\begin{tabular}{@{}l
                S[table-format=2.2] 
                S[table-format=2.2]
                S[table-format=2.2]
                S[table-format=2.2]
                S[table-format=2.2]
                S[table-format=2.2]
                S[table-format=2.2]
                S[table-format=2.2]
                S[table-format=2.2]
                S[table-format=2.2]@{}}
\toprule
 &  \multicolumn{6}{c}{\textbf{HOVER}} & \multicolumn{4}{c}{\textbf{EX-FEVER}} \\
 &  \multicolumn{2}{c}{\textbf{2-hops}} & \multicolumn{2}{c}{\textbf{3-hops}} & 
\multicolumn{2}{c}{\textbf{4-hops}} & \multicolumn{2}{c}{\textbf{2-hops}} &  \multicolumn{2}{c}{\textbf{3-hops}} \\
 \cmidrule[0.4pt](r{0.125em}){2-3}%
 \cmidrule[0.4pt](r{0.125em}){4-5}%
 \cmidrule[0.4pt](r{0.125em}){6-7}%
 \cmidrule[0.4pt](r{0.125em}){8-9}%
 \cmidrule[0.4pt](r{0.125em}){10-11}%
 & F\text{1} & Acc & F\text{1} & Acc & F\text{1} & Acc & F\text{1} & Acc & F\text{1} & Acc \\ 
\midrule
RAG & 57.47 & 57.60 & 49.49 & 52.40 & 50.74 & 53.00 & 68.67 & 68.80 & 68.54 & 68.60 \\
Decomp-Verify & 59.67 & 60.80 & 47.13 & 53.71 & 43.59 & 52.80 & 63.09 & 64.60 & 54.76 & 59.00 \\
ProgramFC & 55.41 & 55.60 & 50.39 & 51.20 & 51.15 & 52.80 & 56.60 & 56.80 & 56.31 & 57.60 \\
BiDeV & 59.38 & 62.40 & 50.55 & 56.20 & 42.09 & 52.60 & 58.73 & 62.20 & 51.60 & 57.60 \\
QACheck & 55.90 & 56.20 & 48.08 & 50.20 & 51.07 & 52.80 & 55.84 & 56.20 & 60.14 & 60.40 \\
FOLK & 61.08 & 61.40 & 59.07 & 59.20 & 57.65 & 58.20 & 80.12 & 80.20 & 73.93 & 74.20 \\
\midrule
\textbf{HARIS} & \textbf{69.31} & \textbf{69.40} & \textbf{62.33} & \textbf{62.80} & \textbf{59.84} & \textbf{61.00} & \textbf{80.12} & \textbf{80.20} & \textbf{73.93} & \textbf{74.20} \\
\bottomrule
\end{tabular}
    \end{threeparttable}
\caption{Direct performance comparison of different methods (Qwen3-4B based) with HARIS.\protect\footnotemark}
\label{tab:qwen3_main}
\end{table*}
\footnotetext{For Search-o1, its official implementation is specifically designed for QwQ reasoning models and is not directly configurable with Qwen3 models.}

\subsection{Training Settings}
\label{appendix:training_settings}
\subsubsection{LLM-as-a-Judge}
\label{appendix:llm_as_a_judge}
We use GPT-4o-mini as the judge for evaluating the final output usefulness of the search agent. This is implemented using DSPy~\citep{khattab2024dspy}, which allows for customizing signature to define prompt-based LLM classification. The signature used in our experiments is provided in Table~\ref{tab:signature}. The final score is set to 1 if the output `is\_useful' variable contains "yes" and 0 otherwise.

\subsubsection{Hardware \& Hyperparameter}
All experiments, including HARIS and the baselines, were conducted on a server with 4×H20 GPUs and a cluster of 8×A100 nodes. Our implementation is based on the verl framework~\citep{sheng2024hybridflow}. Key hyperparameters include: rollout group size of 5, tensor parallel size (tp) of 2, batch size of 48, temperature of 1.0, learning rate of 1e-6, and KL coefficient of 0.001.

HARIS experiments were mainly run on the 4×H20 server. Retrieval services were hosted on a single GPU using FastAPI. For reasoning agent training, we used vLLM~\citep{kwon2023efficient} to serve the search agent endpoint on one GPU, while the remaining GPUs were allocated to high-level agent training. Two GPUs for training the reasoning agent and one GPU for the search agent inference service. In the single-agent setting, two GPUs were used for training. Due to GPU memory constraints, we set the maximum context length to 8192 tokens.

\section{Additional Experiments}
\label{appendix:add_exp}
\subsection{Direct comparison} 
To enable direct comparison on the same base model, we run the baselines using Qwen3-4B as the base LLM. The results are summarized in Table~\ref{tab:qwen3_main}. As shown, when using the same base LLM, HARIS significantly outperforms the baselines, demonstrating its effectiveness.

\subsection{Supervised Finetuning}
We provide an experiment comparing HARIS with supervised finetuning (SFT). Specifically, in an explainable fact-checking setting, we enabled the thinking mode of Qwen3-4B to generate responses using the same training set as HARIS. To ensure the model learns the correct target sequence, for each claim in the training data, we repeatedly sampled responses until the final prediction was correct. The prompt used can be found in Table~\ref{tab:prompt_templates_2}. Training epochs and learning rates were kept the same as HARIS's setting. The performance results are summarized in Table~\ref{tab:sft_comparison}. Overall, HARIS outperforms supervised fine-tuning across all datasets and hop settings.

\subsection{ReAct \& Model Scaling}
One might be concerned that decoupling the search and reasoning agents primarily compensates for the limitations of smaller models (such as our 4B backbone). However, in our main experiments, the Search-o1 baseline employs a larger 32B model, yet it still underperforms compared to HARIS. To further investigate the effect of model scaling, we implemented a ReAct LLM Agent baseline and conducted experiments using Qwen3-4B, 8B, and 14B. In this setup, the agent performs Wikipedia searches and leverages the retrieved documents as observations. The F1 results are presented in Table~\ref{tab:qwen3_react}. As shown, simply increasing the model size does not always result in substantial performance gains. These results suggest that our multi-agent, decoupled approach offers distinct advantages.

\begin{table}[t]
\centering
\small
\setlength{\abovecaptionskip}{0.2cm}
\setlength{\belowcaptionskip}{-0.3cm}
\begin{tabular}{lcccc}
\toprule
& \multicolumn{2}{c}{\textbf{SFT}} & \multicolumn{2}{c}{\textbf{HARIS}} \\
\cmidrule(lr){2-3} \cmidrule(lr){4-5}
& F1 & Acc & F1 & Acc \\
\midrule
$\text{EX-FEVER}_{\text{2hops}}$ & 71.93 & 72.60 & 80.12 & 80.20 \\
$\text{EX-FEVER}_{\text{3hops}}$  & 63.95 & 66.20 & 73.93 & 74.20 \\
$\text{HOVER}_{\text{2hops}}$  & 57.05	& 59.40 & 69.31 & 69.40 \\
$\text{HOVER}_{\text{3hops}}$  & 42.03 & 51.40 & 62.33 & 62.80 \\
$\text{HOVER}_{\text{4hops}}$  & 41.24 & 52.00 & 59.84 & 61.00 \\
\bottomrule
\end{tabular}
\caption{Performance comparison between supervised finetuning(SFT) and HARIS.}
\label{tab:sft_comparison}
\end{table}

\begin{table}[t]
\centering
\small
\setlength{\abovecaptionskip}{0.2cm}
\setlength{\belowcaptionskip}{-0.3cm}
\begin{tabular}{lccc|c}
\toprule
                & \multicolumn{3}{c|}{\textbf{ReAct}} & \textbf{HARIS} \\
                & \text{4B} & \text{8B} & \text{14B} & \text{4B} \\
\midrule
$\text{EX-FEVER}_{\text{2hops}}$  & 72.17 & 71.56 & 72.58 & 80.12 \\
$\text{EX-FEVER}_{\text{3hops}}$  & 67.94 & 63.91 & 70.91 & 73.93 \\
$\text{HOVER}_{\text{2hops}}$     & 63.59 & 63.19 & 64.57 & 69.31 \\
$\text{HOVER}_{\text{3hops}}$     & 54.94 & 58.19 & 55.76 & 62.33 \\
$\text{HOVER}_{\text{4hops}}$     & 51.24 & 53.17 & 56.02 & 59.84 \\
\hline
\end{tabular}
\caption{F1 performance comparison between HARIS and ReAct agent based on different model sizes.}
\label{tab:qwen3_react}
\end{table}


\section{Prompts \& Examples}
\subsection{Prompt Template}
\label{appendix:prompt_template}
The prompt templates used for reasoning agent and search agent are shown in Table~\ref{tab:prompt_templates}.

\begin{table*}[h]
\centering
\renewcommand{\arraystretch}{1.3}
\begin{tabular}{p{0.9\textwidth}}
\hline
\textbf{Prompt Template for Low-Level Search Agent} \\
\hline
You are a helpful assistant tasked with gathering information to answer a question step by step with the help of the wikipedia search tool. Given a question, you need to think about the reasoning process in the mind and how to gather sufficient information to finally report the gathered information clearly based on the information you have found. Your task includes answering the question and reporting relevant information you have found clearly. During thinking, you can invoke the wikipedia search tool to search for fact information about specific topics if needed. The reasoning process and reported information are enclosed within <think> </think> and <report> </report> tags respectively, 
and the search query and result are enclosed within <search> </search> and <result> </result> tags respectively...\\  
\hline

\textbf{Prompt Template for High-Level Reasoning Agent} \\
\hline
You are a helpful assistant tasked with verifying the truthfulness of a claim step by step, with the support of a Wikipedia search agent. Given a claim, you need to think about the reasoning process in the mind and then provide the verification result (Support or Refute). During thinking, if needed, ask factual questions to the Wikipedia search agent. This is a multi-hop claim verification task, the reasoning may involve identifying intermediate facts (bridging facts) that are not explicitly mentioned in the claim but are necessary to verify its truthfulness. \\
For the wikipedia agent to clearly understand the question, follow these guidelines: \\
1. Begin the question with clear interrogatives. \\
2. Questions must be self-contained—do not refer to "the claim" or use vague pronouns like "it" or "that". \\
3. Avoid context-dependent phrases like "in the claim" or "based on that". \\
The reasoning and questioning process should be interleaved using the following tags: \\
- Use <think> </think> to enclose the reasoning process. \\
- Use <question> </question> to pose a factual question. \\
- The agent will return relevant information inside <result> </result> tags. \\
- The final binary decision—**Support** or **Refute**—must be wrapped in LaTeX format as \textbackslash boxed\{Support\} or \textbackslash boxed\{Refute\} inside the <verification> tag... \\
\hline
\end{tabular}
\caption{Prompt templates for the low-level search agent and high-level reasoning agent.}
\label{tab:prompt_templates}
\end{table*}

\begin{table*}[h]
\centering
\renewcommand{\arraystretch}{1.3}
\begin{tabular}{p{0.9\textwidth}}
\hline
\textbf{Prompt Template for Single Agent} \\
\hline
You are a helpful assistant tasked with verifying the truthfulness of a claim step by step, with the help of the wikipedia search tool. Given a claim, you need to first think about the reasoning process in the mind and then provide the boolean verification result (Support or Refute). During thinking, you can invoke the wikipedia search tool to search for fact information about specific topics if needed. The reasoning process and answer are enclosed within <think> </think> and <answer> </answer> tags respectively, and the search query and result are enclosed within <search> </search> and <result> </result> tags respectively... \\
\hline

\textbf{Prompt Template for Supervised Finetuning} \\
\hline
<|im\_start|>user \\
Given a claim and its retrieved evidence, determine whether the claim is 'Support' or 'Refute'. \\
Claim: {claim} \\
Evidence: {retrieved\_evidence} \\
Wrap your final answer in <answer> and </answer> tags (e.g. <answer>Support</answer> or <answer>Refute</answer>)<|im\_end|> \\
<|im\_start|>assistant \\
<think> \\
\hline
\end{tabular}
\caption{Prompt templates for the single agent and supervised finetuning.}
\label{tab:prompt_templates_2}
\end{table*}

\begin{table*}[ht]
\centering
\renewcommand{\arraystretch}{1.3}
\begin{tabular}{p{0.9\textwidth}}
\hline
\textbf{LLM-as-a-Judge DSPy Signature} \\
\hline
\texttt{class SearchAgentRewardSignature(dspy.Signature):} \\

\quad \texttt{question: str} = dspy.InputField(desc="The question for which information must be gathered") \\

\quad \texttt{ground\_truth\_answer: str} = dspy.InputField(desc="The correct answer to the question") \\

\quad \texttt{gathered\_information: str} = dspy.InputField(desc="Information gathered by the search agent intended to help answer the question") \\

\quad \texttt{is\_useful: Literal["yes", "no"]} = dspy.OutputField(desc="Determine whether the gathered information is sufficient and useful to derive the correct answer") \\
\hline

\textbf{Pseudo Ground-Truth Answer Signature} \\
\hline
\texttt{class PseudoGroundTruthQA(dspy.Signature):} \\

\quad \texttt{claim: str} = dspy.InputField() \\

\quad \texttt{veracity: Literal["true", "false"]} = dspy.InputField(desc="The veracity of the claim") \\

\quad \texttt{evidence: dict[str, list[str]]} = dspy.InputField(desc="Supporting evidence/explanation for the veracity of the claim") \\

\quad \texttt{question: str} = dspy.InputField(desc="A relevant question asked by a fact-checking agent") \\

\quad \texttt{answer: str} = dspy.OutputField(desc="The answer to the question. If no answer is applicable, return 'None'") \\
\hline

\textbf{Classification Signature for RAG and Decompose-Then-Verify} \\
\hline
\texttt{class ClaimVerificationSignature(dspy.Signature):} \\

\quad \texttt{claim: str} = dspy.InputField(desc="The claim to be checked") \\

\quad \texttt{context: str} = dspy.InputField(desc="The retrieved evidence for the claim") \\

\quad \texttt{veracity: Literal['Support', 'Refute']} = dspy.OutputField(desc="Given the claim and the retrieved evidence, determine whether the claim is 'Support' or 'Refute'") \\
\hline

\end{tabular}
\caption{Signature definitions for search agent reward evaluation, search agent training data synthesis, and baselines claim verification.}
\label{tab:signature}
\end{table*}

\subsection{Synthesized Training Data}
Table~\ref{tab:synthesized_data_examples} shows two examples of synthesized training data for the search agent. Example 2 leverages human-annotated explanations as evidence. The synthesis process uses a Chain-of-Thought (CoT) prompting format, with the \texttt{rationale} field capturing GPT-4o's intermediate reasoning before producing the final answer.

\begin{table*}[ht]
\centering
\small
\renewcommand{\arraystretch}{1.3}
\begin{tabular}{|p{0.9\textwidth}|}
\hline
\textbf{Synthesized Raw Training Data Example 1} \\
\hline
\begin{lstlisting}
{
  "input": "94 episodes were in the original run of the HBO show in which the voice actress of Miss Spider played the character Charlotte York Goldenblatt.",
  "evidence": "Miss Flora Spider is the title character in the bestselling children's books by David Kirk. She originally appeared in cartoon format in the movie \"Miss Spider's Sunny Patch Kids\" (with Brooke Shields providing Miss Spider's voice), which spun off the \"Miss Spider's Sunny Patch Friends\" series (with Kristin Davis providing her voice this time around), and \"Miss Spider's Sunny Patch Friends - Harvest Time Hop and Fly\" video game...",
  "explanation": null,
  "label": true,
  "question": "Who is the voice actress for the character \"Miss Spider\"?",
  "qa_response": {
    "rationale": "To produce the answer. We start by identifying the character \"Miss Spider\" and the voice actress associated with her. According to the evidence, Miss Spider is a character in the children's books by David Kirk and has appeared in various media formats. The evidence states that Brooke Shields provided Miss Spider's voice in the movie \"Miss Spider's Sunny Patch Kids,\" while Kristin Davis provided her voice in the \"Miss Spider's Sunny Patch Friends\" series. Since the question asks for the voice actress for the character \"Miss Spider,\" we need to determine which context is being referred to. However, given the context of the claim and the evidence provided, it is most relevant to identify Kristin Davis as the voice actress for the series, as she is also the actress who played Charlotte York Goldenblatt in \"Sex and the City.\" Therefore, the answer to the question is Kristin Davis.",
    "answer": "Kristin Davis"
  }
}
\end{lstlisting} \\
\hline
\textbf{Synthesized Raw Training Data Example 2} \\
\hline
\begin{lstlisting}
{
  "input": "Journey is an American rock band formed in 1974 by former members of band such as an American rock band formed in 1966.",
  "evidence": null,
  "explanation": "Journey is formed in 1973 by former members of band such as Steve Miller Band, not 1974. The Steve Miller Band is an American rock band formed in 1966.",
  "label": false,
  "question": "What American rock band was formed in 1966 and had members who later joined Journey?",
  "qa_response": {
    "rationale": "To produce the answer. We start by identifying the American rock band formed in 1966. According to the evidence, the Steve Miller Band was formed in 1966. Next, we need to determine if any members from this band later joined Journey. The evidence states that Journey was formed by former members of bands such as the Steve Miller Band. Therefore, the Steve Miller Band is the band in question.",
    "answer": "The Steve Miller Band."
  }
}
\end{lstlisting} \\
\hline
\end{tabular}
\caption{Synthesized raw data examples for training the reasoning and search agents.}
\label{tab:synthesized_data_examples}
\end{table*}

\begin{figure*}[h]
\centering
\includegraphics[width=1.\textwidth]{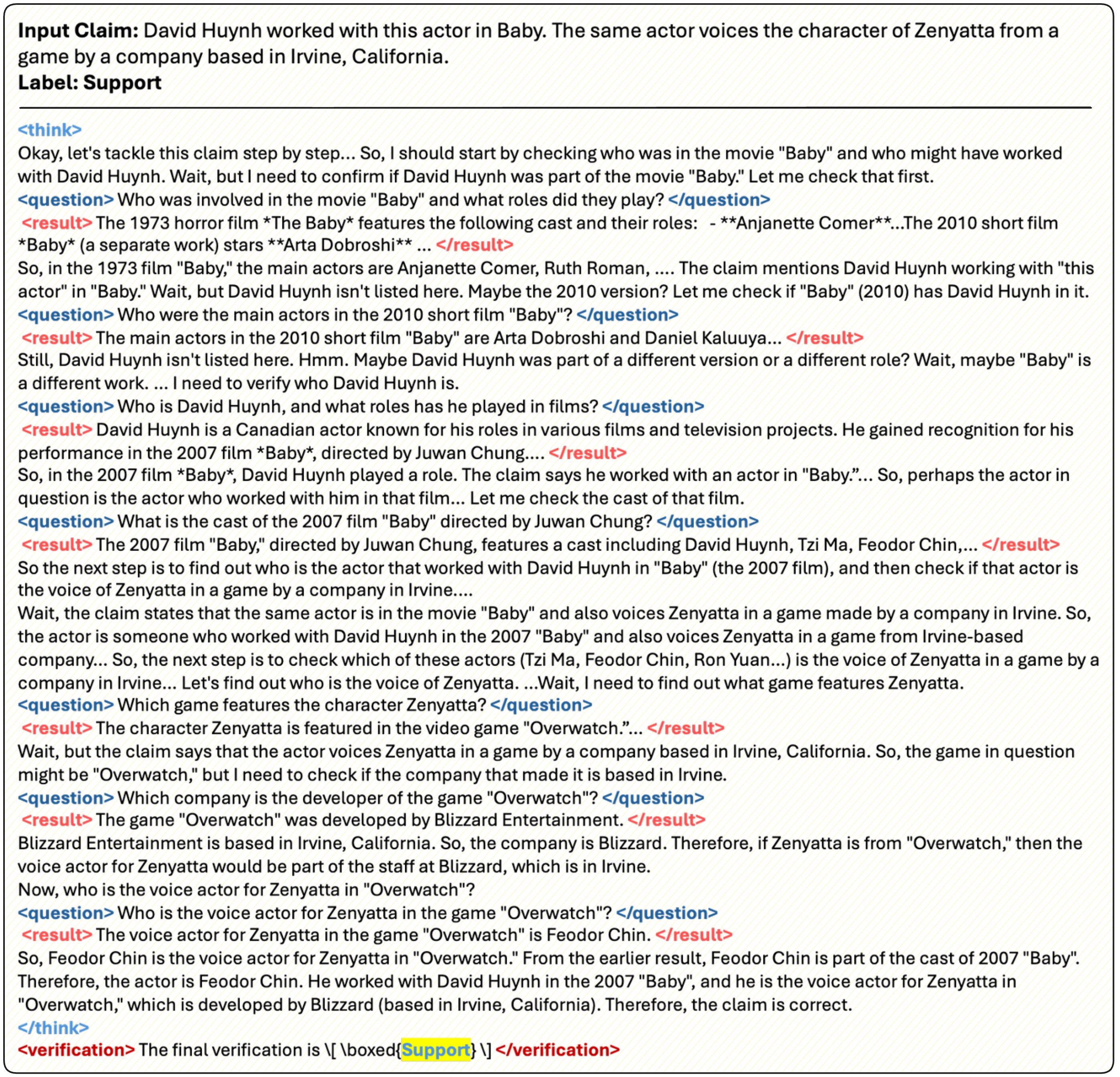}
\caption{Different rollout example from reasoning agent for a complex multi-hop claim. } 
\label{fig:case_study_2}
\end{figure*}

\subsection{Synthesized Training Data}
Table~\ref{tab:synthesized_data_examples} shows two examples of synthesized training data for the search agent. Example 2 leverages human-annotated explanations as evidence. The synthesis process uses a Chain-of-Thought (CoT) prompting format, with the \texttt{rationale} field capturing GPT-4o's intermediate reasoning before producing the final answer.

\subsection{Single-Agent \& Multi-Agent Cases}
\label{appendix:single_case_study}
\begin{figure*}[h]
\centering
\includegraphics[width=1.\textwidth]{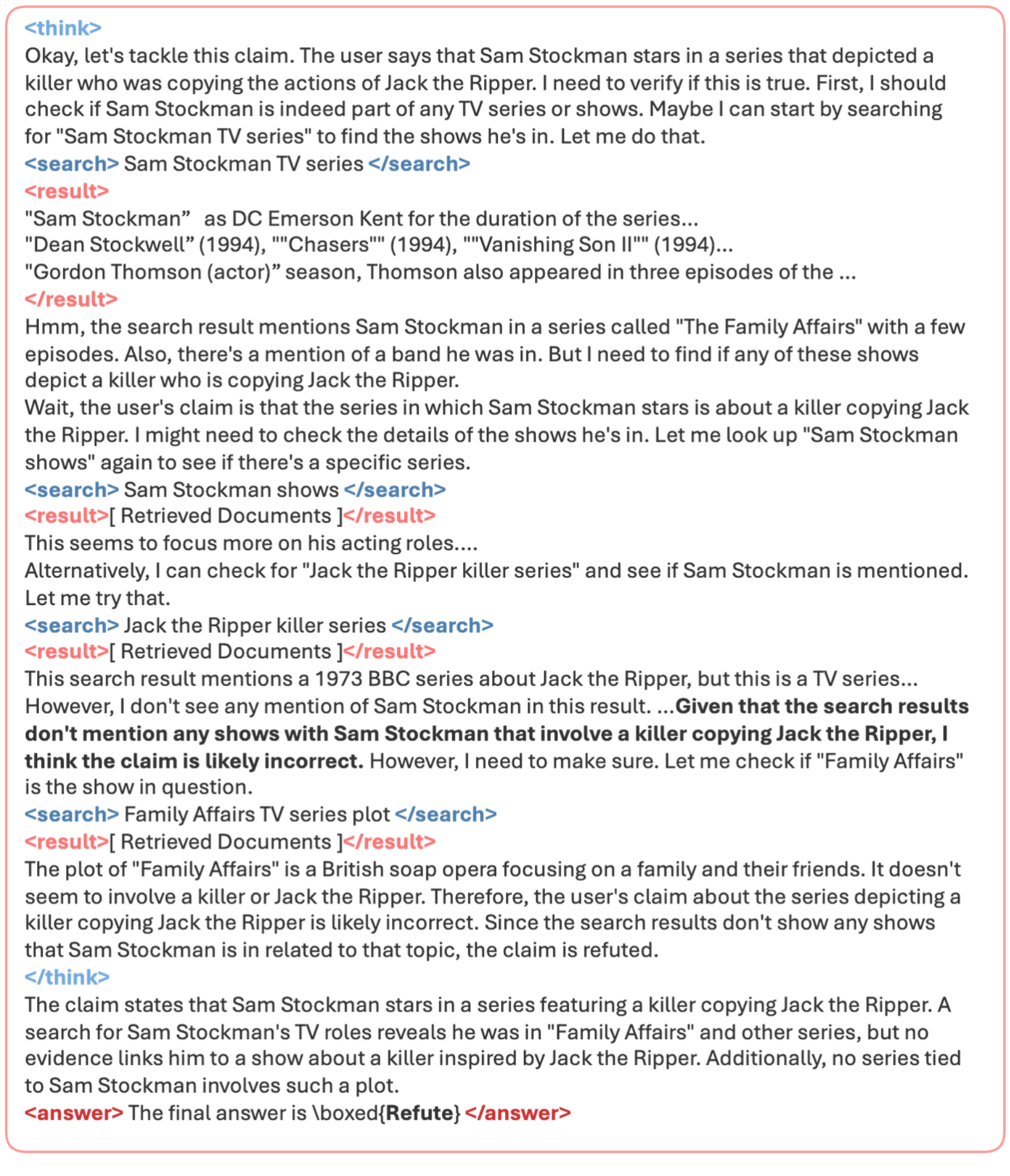}
\caption{Example of a trained single agent rollout. The input claim `Sam Stockman stars in a series that depicted a killer. The killer was copying the actions of Jack the Ripper.' is correct. The agent issues several queries but fails to retrieve sufficient evidence. Without a clear separation of roles, it struggles to iteratively refine its search direction and prematurely converges on partial evidence, ultimately leading to an incorrect verification. } 
\label{fig:single_agent_exp}
\end{figure*}

\begin{figure*}[h]
\centering
\includegraphics[width=1.\textwidth]{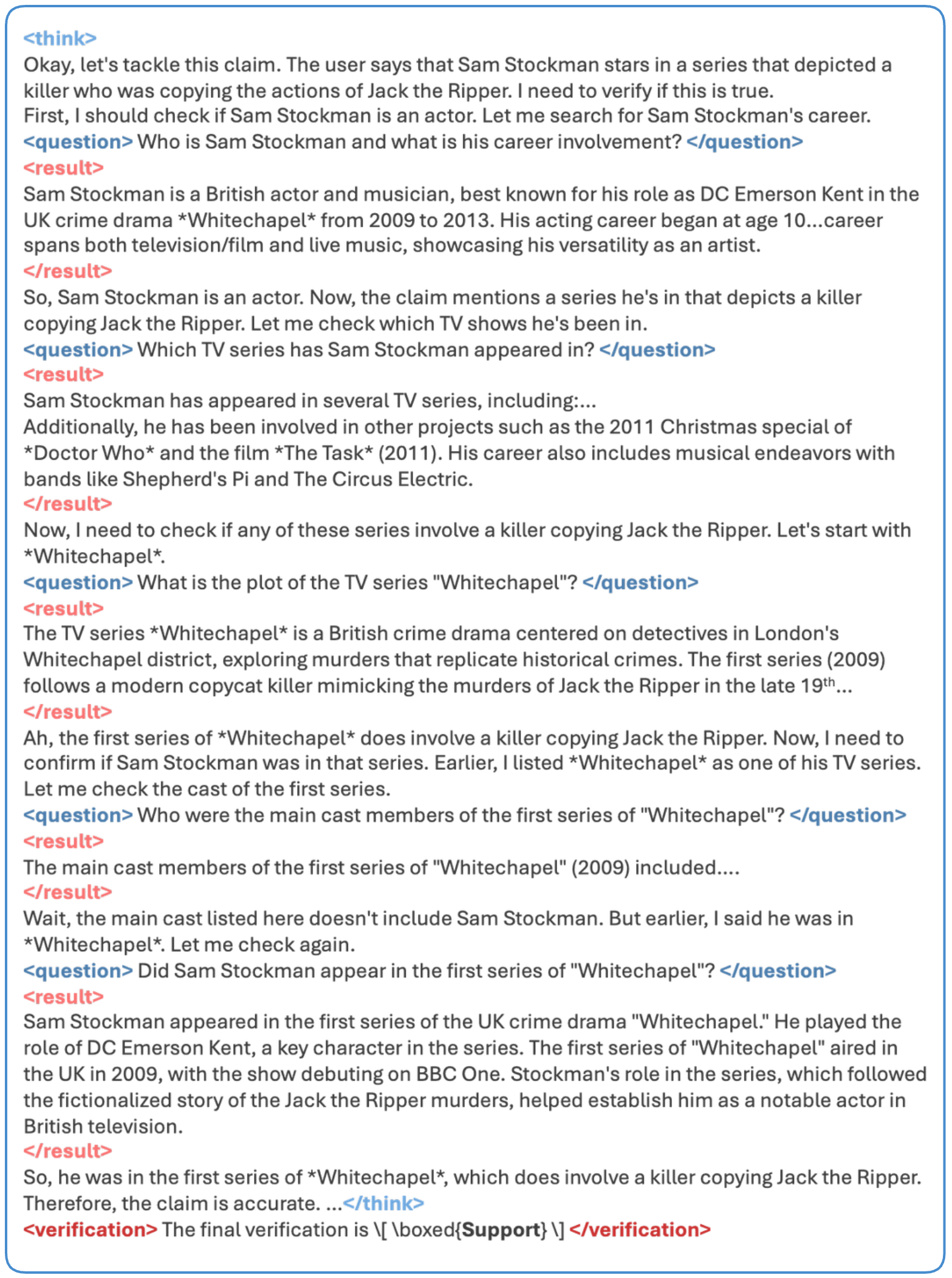}
\caption{Example of HARIS's reasoning agent rollout. The input claim `Sam Stockman stars in a series that depicted a killer. The killer was copying the actions of Jack the Ripper.' is correct. The reasoning and search agents coordinate effectively: the reasoning agent identifies uncertain links and delegates targeted queries, while the search agent retrieves precise evidence, enabling correct verification. } 
\label{fig:multi_agent_exp}
\end{figure*}

To better understand the behavioral differences between single-agent and coordinated reasoning-search approaches, we compare two rollouts for the same claim in Figures~\ref{fig:single_agent_exp} and~\ref{fig:multi_agent_exp}.

In the single-agent case, the model issues several searches but fails to effectively refine its queries. For each aspect it explores, it stops short of deeper investigation and prematurely converges on partial evidence. With reasoning and retrieval entangled in a single generation loop, the agent lacks feedback mechanisms to reassess or adjust its direction, ultimately producing an incorrect verification.

In contrast, HARIS decouples reasoning and search into specialized agents. The reasoning agent identifies uncertain links and formulates precise questions, while the search agent iteratively gathers relevant evidence to support or refute each hypothesis. This coordinated process enables effective disambiguation, deeper exploration, and accurate multi-hop reasoning. The comparison highlights how HARIS’s multi-agent design leads to more robust, interpretable verification under ambiguity and incomplete evidence.

\section{Human Evaluation Details}
\label{appendix:human_eval}
\begin{figure*}[h]
\centering
\includegraphics[width=1.\textwidth]{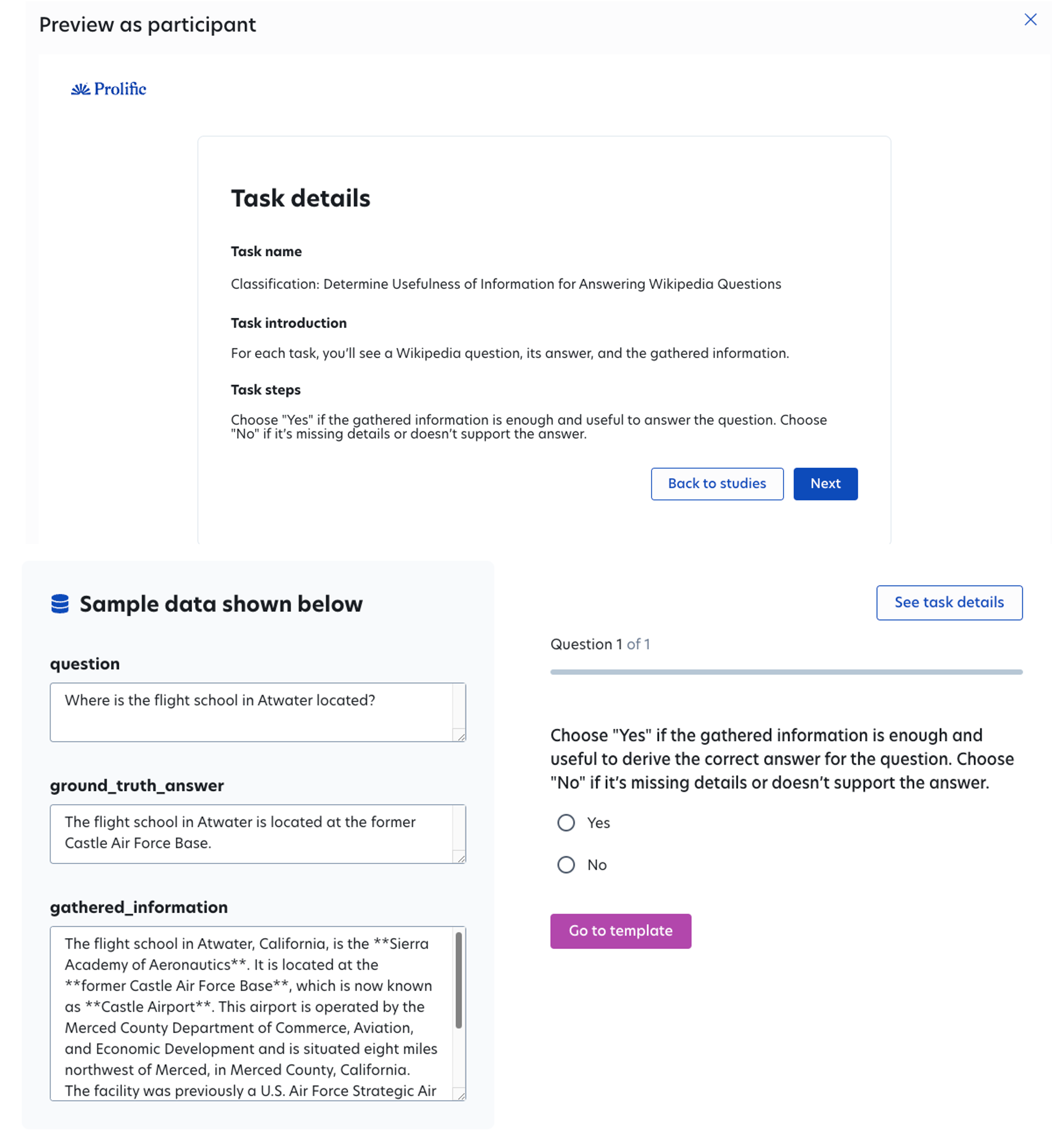}
\caption{Preview of the human evaluation panel.} 
\label{fig:human_eva_panel}
\end{figure*}

As described in Section~\ref{sec:search_agent_reward}, we conducted a human evaluation to assess the reliability of our LLM-as-a-Judge setup. We sampled 150 questions from a held-out set of synthesized QA data, using the trained HARIS search agent to gather information for each. Two annotators from the Prolific platform\footnote{\url{https://www.prolific.com/}}, each paid £20, independently evaluated 75 responses following the same guidelines as the LLM-as-a-Judge. They judged whether the retrieved information was sufficient and useful for deriving the correct answer. The results showed a Cohen’s Kappa of 0.81 and a 93.3\% agreement rate, indicating strong consistency. These findings confirm that our LLM-as-a-Judge metric closely aligns with human judgments. An example of the annotation panel is shown in Figure~\ref{fig:human_eva_panel}. For data consent, we selected the AI task annotation category on the platform, and annotators were informed that the collected data would be used to evaluate LLM outputs.

\end{document}